\documentclass[11pt,a4paper]{article}
\usepackage[hyperref]{emnlp2020}
\usepackage{times}
\usepackage{latexsym}

\usepackage{microtype}
\usepackage{xspace,mfirstuc,tabulary}
\usepackage{times}
\usepackage{latexsym}
 \usepackage{tabu,array,multirow,graphicx}
 \usepackage{float}
\usepackage{url}

\usepackage{graphicx}
\usepackage{xcolor}
\usepackage{pgfplots}
\usepackage{amsmath}
\usepackage{amssymb}
\usepackage{tikz}
\usepackage{booktabs}
\usepackage{subcaption}
\usepackage{todonotes}

\definecolor{bblue}{HTML}{4F81BD}
\definecolor{rred}{HTML}{C0504D}
\definecolor{ggreen}{HTML}{9BBB59}
\definecolor{ppurple}{HTML}{9F4C7C}
\def\aclpaperid{}

\title{Improved Semantic Role Labeling using Parameterized Neighborhood Memory Adaptation}

\author{Ishan Jindal\textsuperscript{a},  Ranit Aharonov\textsuperscript{a}, Siddhartha Brahma\textsuperscript{b}{\thanks{\textsuperscript{b} Work done while at IBM Research}}, Huaiyu Zhu\textsuperscript{a}, Yunyao Li\textsuperscript{a} \\
\textsuperscript{a}IBM Research, Almaden Research Center, CA 95120 \\
\textsuperscript{b}Google Research, Berlin, Germany\\ 
  {\tt \{ishan.jindal,ranit.aharonov2\}@ibm.com,}\\ {\tt\{huaiyu, yunyaoli\}@us.ibm.com},\\{\tt sidbrahma@gmail.com}}
  
\date{}

\aclfinalcopy % Uncomment this line for the final submission

\begin{document}

\maketitle

\begin{abstract}
Deep neural models achieve some of the best results for semantic role labeling. Inspired by instance-based learning that utilizes nearest neighbors to handle low-frequency context-specific training samples, we investigate the use of memory adaptation techniques in deep neural models. We propose a parameterized neighborhood memory adaptive (PNMA) method that uses a parameterized representation of the nearest neighbors of tokens in a memory of activations and makes predictions based on the most similar samples in the training data. We empirically show that PNMA consistently improves the SRL performance of the base model irrespective of types of word embeddings. Coupled with contextualized word embeddings derived from BERT, PNMA improves over existing models for both span and dependency semantic parsing datasets, especially on out-of-domain text, reaching F1 scores of $80.2\%$, and $84.97\%$ on CoNLL2005, and CoNLL2009 datasets, respectively.
\end{abstract}

\section{Introduction}

Semantic role labeling (SRL) is the task of identifying predicate-argument structures from a given sentence based on semantic frames and their roles~\cite{Jurafsky2008SpeechAL}. It is a fundamental task for natural language understanding.

\begin{figure}[!t]
    \centering
    \includegraphics[width=0.9\columnwidth]{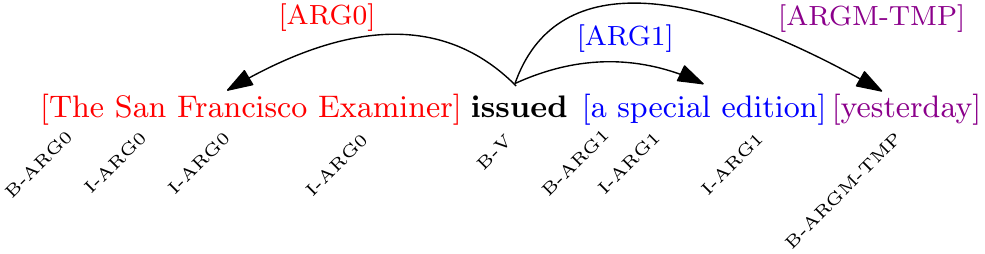}
    \caption{Semantic Role Labeling with BIO tags}
    \label{fig:srl_example}
    % \vspace{-1.2em}
\end{figure}

As shown in Fig.\ref{fig:srl_example}, SRL is typically decomposed into four sub-tasks:
1) \emph{predicate identification (e.g. issue)}; 
2) \emph{sense disambiguation (e.g. issue.01)}; 3) \emph{argument identification} for each predicate \emph{(e.g. a special edition)}; and 4) \emph{role classification} of identified arguments (\emph{e.g. ARG1}). A popular approach for SRL \cite{li2018unified} 
is to assume predicates are given and convert the argument prediction problem into a sequence tagging problem both for span and dependency type arguments.

Deep neural models achieve some of the best results in predicting semantic roles over standard benchmark datasets \cite{He2017DeepSR,Tan2018DeepSR,Ouchi2018ASS,li2019dependency,kasai2019syntax}.%\todo{Yunyao: Need more recent citations}
Meanwhile, a very simple instance-based learning method, K-SRL \cite{Akbik2016KSRLIL} has been shown strong performance using a distance function based on syntactic features. The semantic role of a candidate test word is determined by taking a majority vote of the labels of closest words present in a memory populated by words from the training set. 

A natural question is whether one can apply such memory based methods with features derived from deep neural models. Memory based and memory adaptive learning methods have been successfully applied to language modeling and other problems e.g. the continuous cache model \cite{Grave2016ImprovingNL,Grave2017UnboundedCM}. The memory contains activations (e.g. from a RNN) observed during training and the associated labels.  
\citet{Sprechmann2018MemorybasedPA} propose an inference time memory based parameter adaptation method for language modeling. At each point during inference, they use its contextual representation to compute $K$ nearest neighbors in the memory. The labels of the neighbors are then used to update the classification layer for final prediction. 

In this paper, we propose a \emph{parameterized neighborhood memory adaptive} (PNMA) method to exploit the information contained in the memory of activations. Our method proceeds in two phases. 
% \vspace{-\topsep}
\begin{itemize}
\item \textbf{Memory Generation:} We populate a memory consists of each token representations from the training set and compute the $K$ nearest neighbors for each token using Euclidean distance. We hypothesize and show empirically that the nearest neighbors contain valuable information about the correct label of $w$ even when the base model prediction itself is wrong.

\item \textbf{PNMA:} We then exploit the memory constructed earlier by computing a \emph{parameterized} single vector representation $\mathbf{n}_K(w) \in \mathbb{R}^d$ of the $K$ nearest neighbors, which is then used to retrain the classification layers of the base model. Instead of a single representation $\mathbf{h}(w)$, the nearest neighbors define an ensemble of $K$ representations close to $\mathbf{h}(w)$, which we combine using learned parameters into  $\mathbf{n}_K(w)$.

\end{itemize}
%We apply PNMA to both span style (CoNLL2005 and CoNLL2012) and dependency style (CoNLL2009) SRL benchmark datasets. 
We show empirically that PNMA improves the performance of the base SRL model, both without and with the use of pretrained word embeddings. We further demonstrate that PNMA addresses the low-frequency predicates in the training data and improves the base model prediction on low-frequency exceptions. Our best model combining PNMA and BERT achieves new state-of-the-art F1 scores on both datasets (for a single model), 
%with a 2.0(4.0) point increase over existing methods on the out-of-domain Brown test set in CoNLL2005(CoNLL2009).
with a 2.0 point increase over existing methods on the out-of-domain Brown test set in CoNLL2005 and 5.0 point increase on CoNLL2009.

In the rest of the paper, we describe the PNMA model in Sec.~\ref{sec:pnma} and demonstrate its efficacy with extensive empirical evaluation in Sec.~\ref{sec:results} and analysis in Sec. \ref{sec:analysis}. We then review the existing literature in Sec.~\ref{sec:review} and conclude in Sec.~\ref{sec:conc}. 

\begin{figure*}[!t]
    \centering=
    \includegraphics[width=0.8\textwidth]{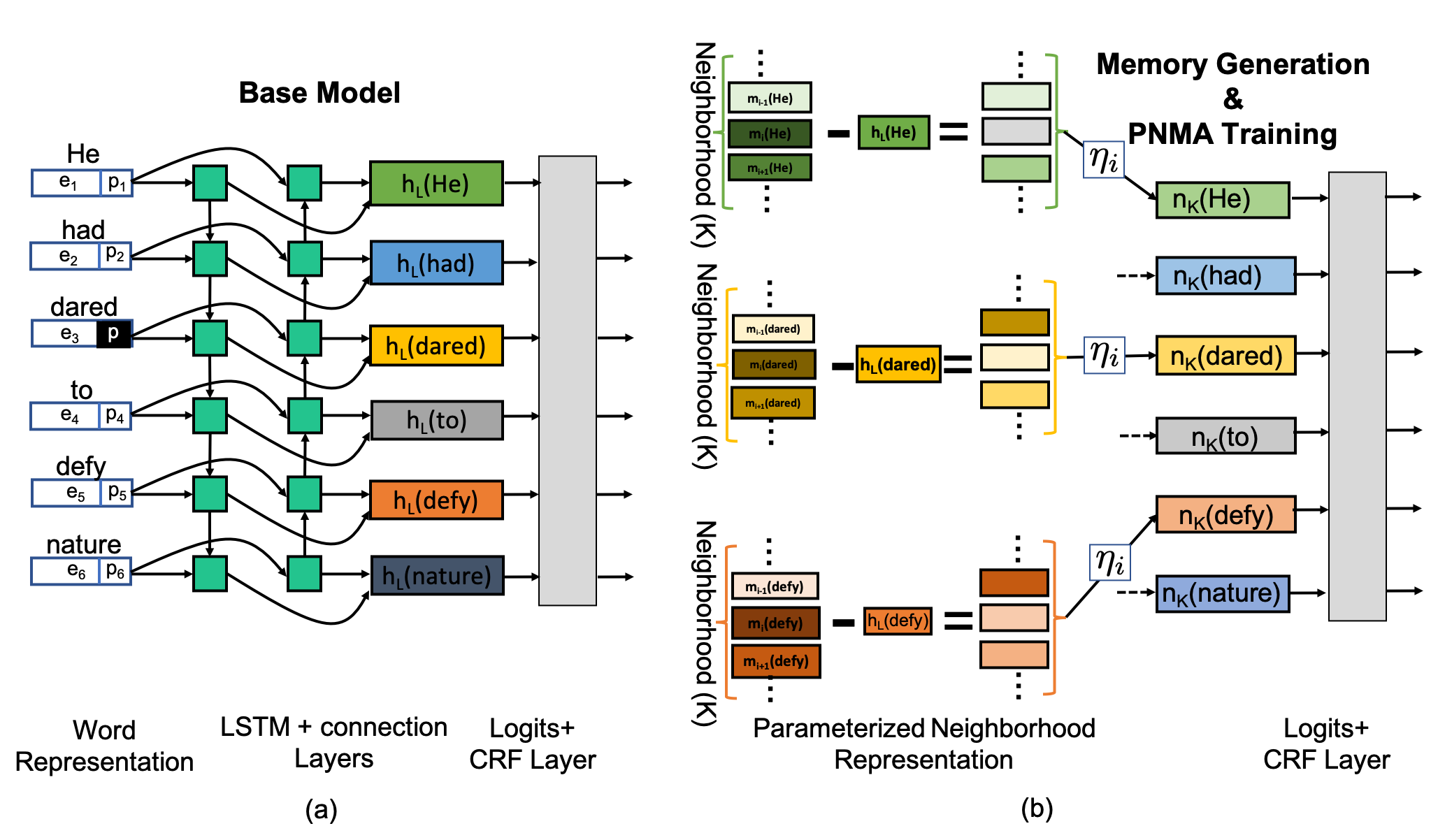}
    \caption{(a) Base model architecture for SRL.
    (b) Learning the parametrized neighborhood representation, $\mathbf{n}_K(w)$.
    When learning $\mathbf{n}_K(w)$, the word representation and LSTM + connection layers are kept fixed.}
    \label{fig:basemodel}
    % \vspace{-1em}
\end{figure*}
\section{Model}
\label{sec:pnma}
We formulate SRL as a sequence tagging problem both for span and dependency type arguments~\cite{Ouchi2018ASS}. For datasets with span style arguments, we convert the semantic role to BIO tags. We assume that the input to the model is a sentence $S$ with $n$ tokens. 
Knowing the predicate position in a sentence is previously known to improve the performance of argument classification \cite{li2018unified}. Since all benchmark datasets have the marked predicate position, we use this information and assume that each token is tagged with a 0/1 bit indicating the predicate word. 
Our proposed method takes the base model representation for populating the memory.   

\subsection{Base Model for SRL}
We use an Alternating LSTM~\cite{He2017DeepSR,Ouchi2018ASS} that has been successfully used for SRL as the base model. As shown in Fig. \ref{fig:basemodel}, the base model has the following main components.

\noindent\textbf{Word Embeddings:} Word embeddings are the numerical vector representations of the textual words.  Contextualized  word embeddings derived from pretrained language models such as ELMo \cite{Peters2018DeepCW} and BERT \cite{Devlin2018BERTPO} have been shown to give superior performance in many NLP tasks. Here, we denote the embeddings of the $n$ tokens by $\mathbf{e}_1,\cdots,\mathbf{e}_n$.

\noindent\textbf{Predicate Embeddings:} 
We use the marked predicate position to obtain a predicate-specific word representation. We map the predicate indicator bits to dense embeddings $\mathbf{p}_1,\cdots,\mathbf{p}_n$. The combined embedding for token $i$ is  the concatenation $\mathbf{x}_1^i = [\mathbf{e}_i;\mathbf{p}_i]$, which is the input to the LSTM.

\noindent\textbf{LSTM Layers:} To sequentially process the word representations from the previous step, we use an Alternating LSTM architecture, where forward and backward LSTM layers alternate. We use a total of $L=4$ layers. The inputs $\mathbf{x}_l^{1},\cdots, \mathbf{x}_l^{n}$ to the $l$-th layer produces the output $\mathbf{h}_l^1,\cdots,\mathbf{h}_l^n$.

\noindent\textbf{Connection Layers.} To facilitate the flow of gradients between LSTM layers, the input to the $(l+1)$th LSTM layer is 
% \begin{equation*}
%     \vspace{-1em}
    $\mathbf{x}_{l+1}^{i}~=~\text{ReLU}(\mathbf{W}_l[\mathbf{h}_l^i;\mathbf{x}_l^i]) \in~\mathbb{R}^d$, 
%         \vspace{-1em}
% \end{equation*}
where $\mathbf{W}_l$ is a parameter matrix.

\noindent\textbf{CRF Layer:} The representations $\mathbf{h}_L^1,\cdots,\mathbf{h}_L^n~\in~\mathbb{R}^d$ from the LSTM layers are passed onto a CRF layer \cite{Zhou2015EndtoendLO}, which produces predictions for each token.
% as having a particular label from the BIO tagset. 
The CRF layer is trained end-to-end with the other layers.

\subsection{Parameterized Neighborhood Memory Adaptive (PNMA) Model}
After we train a strong multilayer LSTM base model on the SRL dataset, in the first phase we generate the memory of activations and in the second phase we compute the parameterized vector representations of each token and retrain the classification layers of base model. 
Here we describe the two phases of our proposed model in detail. 
\subsubsection{Phase 1: Memory Generation}
\begin{figure}
% \vspace{-0.1in}
\centering
\begin{tikzpicture}
\begin{axis}[
height=0.5\columnwidth,
width=0.75\columnwidth,
xtick={1,4,8,16,32,64},
xticklabel style={},
xlabel={Rank},
y tick label style={/pgf/number format/1000 sep=},
extra y tick style={grid=major, tick label style={xshift=-1cm}},
ylabel={Normalized freqeuncy},
ylabel near ticks,
label style={font=\small},
tick label style={font=\small},
grid=major,
]
\addplot [color=red,mark=*,mark size=1pt] table[x=rank,y=prob] {rank.dat};
\end{axis}
\end{tikzpicture}
% \vspace{-0.1in}
 \caption{Distribution of rank of nearest token with correct label among $K=64$ nearest neighbors.}
    \label{fig:pnma_evidence}
    % \vspace{-1em}
\end{figure}
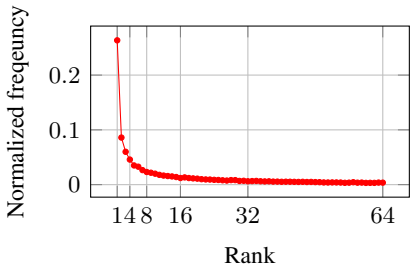

\noindent{\textbf{Intuition:}} We hypothesize that the $K$ nearest neighbors contain valuable information  about the correct label of $w$ even when $\mathbf{h}_L(w)$ itself leads to an incorrect prediction. To test this hypothesis, we train a base SRL model on the CoNLL2005 training set (see details in Section \ref{sec:results}) and compute the predicted labels for the sentences in the validation set.
For each token $w'$ with a wrong predicted label, we compute its $K=64$ nearest neighbors and find the rank (in order of increasing distance) of the first token with the correct label (of $w'$) among the neighbors. The distribution of these ranks, as shown in Fig.~\ref{fig:pnma_evidence}, is highly left skewed. Almost all tokens labeled incorrectly by the base model have a close neighbor in memory that has the correct label. The corresponding distribution for tokens correctly labeled by the base model is even more left skewed. However, it is not clear a-priori which neighbor of $w'$ should we trust. This observation motivates the second phase of training.

\noindent{\textbf{Memory Generation:}} We use the trained base model to populate a memory $M$ with activations $\mathbf{h}_L(w)$ produced by the final LSTM layer on words $w$ in the training set, where
\vspace{-1em}
\begin{equation*}
   M = \{\mathbf{h}_L(w) \text{ where } w \text{ is in the training set.} \}
   \vspace{-1em}
\end{equation*}

The size of $M$ can be varied and our particular choice is discussed in Section \ref{sec:results}. Now, for each token $w$ in the training set, we compute its $K$ nearest neighbors using the Euclidean distance between $\mathbf{h}_L(w)$  and $\mathbf{m}\in M$. We denote the representation of the $i$-th nearest neighbor of $w$ by $\mathbf{m}_i(w)\in  \mathbb{R}^d$ and the set of nearest neighbor vectors by $N_K(w)$.

\noindent{\textbf{Parametrized Representation:}} To exploit the information contained in the neighborhood, we compute a \emph{parameterized} compact representation of $N_K(w)$.  Let $\{~\mathbf{n}_i~\in~\mathbb{R}^d, \text{ for } 1\leq i \leq K\}$ denote a set of \emph{neighborhood} parameters learned during training. Consider the following quantities. 
\vspace{-0.65em}
\begin{equation}
\{\eta_i(w)\}_{i=1}^K = \text{Softmax}\{\mathbf{n}_i\cdot|\mathbf{m}_i(w)-\mathbf{h}_L(w)|\}_{i=1}^K 
\vspace{-0.25em}
\end{equation}
Intuitively, $|\mathbf{m}_i(w)-\mathbf{h}_L(w)|$ (the element-wise absolute difference) is a dense representation of the separation between $w$  and its $i$-th nearest neighbor. The inner product with $\mathbf{n}_i$ followed by the softmax computes a  distribution over the neighbors $\{\eta_i\}_{i=1}^K$, which is then used to compute a compact weighted representation of $N_K(w)$.
    % \vspace{-1em}
\begin{equation}
    \mathbf{n}_K(w) = \sum_{i=1}^{K} \eta_i(w) \mathbf{m}_i(w) \in \mathbb{R}^d
    \label{eq:nbr}
        % \vspace{-1em}
\end{equation}
It is worthwhile to contrast $\mathbf{h}_L(w)$ with $\mathbf{n}_K(w)$. The former is the representation of $w$ produced by the base model, while the latter is derived from an ensemble of $K$ representations that are close to $\mathbf{h}_L(w)$ in $M$. By making $\mathbf{n}_K(w)$ a function of $|\mathbf{m}_i(w)-\mathbf{h}_L(w)|$ and learnable parameters, we aim to extract the most relevant information in $N_K(w)$. We note that we observed consistently better performance by using a distinct $\mathbf{n}_i$ for each neighbor instead of a single vector for all neighbors.

\subsection{PNMA Training}

In the second training phase, we freeze the parameters associated with the LSTM, connection and embedding layers in the base model and update the neighborhood parameters $\{\mathbf{n}_i\}_{i=1}^K$ and the parameters in the classification and CRF layers using $\mathbf{n}_K(w)$ and the label of $w$ in the training set. We did not observe benefits by using $\mathbf{h}_L(w)$ itself in the  second training phase.  
At test time, we obtain the SRL label predictions by computing $\mathbf{n}_K(w)$ for each token $w$ in the test sentence, which itself requires the computation of $N_K(w)$ using the LSTM representations of $w$ and $M$.

\begin{table*}[!t]
\resizebox{\textwidth}{!}{
\def\arraystretch{1.}
\setlength{\tabcolsep}{4pt}
\begin{tabular}{llllllllll  || lll  lll }
\toprule
 & \multicolumn{9}{c||}{\textbf{CoNLL2005}} &  \multicolumn{6}{c}{\textbf{CoNLL2012}}\\
\cmidrule(lr){2-10} \cmidrule(lr){11-16}
 & \multicolumn{3}{c}{Validation} & \multicolumn{3}{c}{Test WSJ } &  \multicolumn{3}{c||}{Test Brown} &
\multicolumn{3}{c}{Validation} & \multicolumn{3}{c}{Test} 
\\
% & \multicolumn{3}{l|}{} & \multicolumn{3}{l|}{InDomain} &  \multicolumn{3}{l||}{OODomain} & \multicolumn{3}{l}{} & \multicolumn{3}{|l}{} \\
\cmidrule(lr){2-4} \cmidrule(lr){5-7} \cmidrule(lr){8-10}\cmidrule(lr){11-13} \cmidrule(lr){14-16}
  Model & P & R & F1 &  P & R & F1 & P & R & F1 & P & R & F1 & P & R & F1 \\ \midrule  \midrule
 % \cmidrule(lr){1-1} \cmidrule(lr){2-4} \cmidrule(lr){5-7} \cmidrule(lr){8-10}\cmidrule(lr){11-13} \cmidrule(lr){14-16}
\multicolumn{10}{l}{Syntax Aware Models}\\  \cmidrule(lr){1-1} 
\cite{zhou2019parsing}+ELMo & - & - & - & 87.76 &88.29 & 88.02 & 79.59&  78.64  & 79.11 &  - & - & - & - & - & - \\
\cite{zhou2019parsing}+BERT & - & - & - & 89.04 &88.79 &88.91 & 81.89 &80.98 &81.43 &  - & - & - & - & - & - \\
% \cite{zhou2019parsing}+XLNet & - & - & - & 89.89 &89.74 &89.81  & \underline{\textbf{85.35}} &\underline{\textbf{84.57}} &\underline{\textbf{84.96}} &  - & - & - & - & - & - \\
\cite{li2019dependency}+ELMo & - & - & - & 89.6 &90.1 &89.8 &82.4  &82.6& 82.5 &  - & - & - & - & - & - \\
\cite{li2019dependency}+BERT & - & - & - & \underline{\textbf{90.2}} &\underline{\textbf{91.8}} &\underline{\textbf{91.0}} &\underline{\textbf{83.2}} &\underline{\textbf{84.0}} &\underline{\textbf{83.6}} &  - & - & - & - & - & - \\
Pred-Span-ELMo & - & - & - & - & - & 87.4 & - & - & 80.4 &  - & - & - & - & - & 85.5 \\ 
\midrule \midrule
 \multicolumn{10}{l}{Syntax Agnostic Models}\\ \cmidrule(lr){1-1} 
ALC-Span-ELMo & \underline{\textbf{87.4}} & 86.3 & \textbf{86.9}& \textbf{88.2} & 87.0 & \textbf{87.6} & \textbf{79.9} & 77.5 & 78.7 & \underline{\textbf{87.2}} &  85.5  & \textbf{86.3} & \textbf{87.1} &  85.3 & \textbf{86.2}\\
ALC-BIO-ELMo  & 86.6 & \textbf{86.8} & 86.7 & 87.4 & \textbf{87.3} & 87.3 & 78.5 & \textbf{78.3} & \textbf{78.4} & 86.1 &  \textbf{85.8} &  85.9 &  86.0 &  85.7 &  85.9\\
%\cite{Ouchi2018ASS} & & & & & & & & &\\
% Pred-Span-ELMo & - & - & - & - & - & 87.4 & - & - & \underline{\textbf{80.4}} &  - & - & - & - & - & 85.5 \\ 
% \midrule  \midrule
\multicolumn{10}{l}{Ours (All use BIO)} \\ \cmidrule(lr){1-1} 
Base-Rand             & 79.4 & 79.0 & 79.2 & 80.9 & 80.3 & 80.6 & 69.6 & 67.3 & 68.5 & 79.7 & 77.7  & 78.7 &  77.5 & 77.8  & 78.7 \\
Base-Rand+PNMA & 80.2 & 79.3 & 79.7 & 81.8 & 80.6 & 81.2 & 70.8 & 68.1 & 69.4 & 80.5 & 78.7  & 79.6 &  78.3 & 78.6  & 79.5 \\

Base-ELMo             & 86.4 & 86.4 & 86.4 & 87.6 & 87.2 & 87.4 & 79.1 & 78.1 & 78.6 & 84.1 & 84.9  & 84.5 &  83.7 & 84.9  & 84.3 \\
Base-ELMo+PNMA & 86.9 & 87.0 & 86.9 & 87.6 & 87.3 & 87.5 & 79.8 & 78.9 & 79.3 & 85.0 & 84.9  & 84.9 &  84.9 & 85.1  & 85.0 \\

Base-BERT & 86.9  & 86.9 & 86.9&87.6 &  87.7 & 87.7 & 79.8   &80.0   &79.9 &         85.5 & 86.2  & 85.9 &    85.4 &  86.3 & 85.9 \\
Base-BERT+PNMA &\textbf{87.0} & \underline{\textbf{87.2}}
& \underline{\textbf{87.1}}&\underline{\textbf{88.7}}  & \underline{\textbf{88.0}} &  \underline{\textbf{87.9}}& \underline{\textbf{80.3}}   &\underline{\textbf{80.1}}  & \underline{\textbf{80.2}}& \textbf{86.4} & \underline{\textbf{86.8}} & \underline{\textbf{86.6}}   &  \underline{\textbf{86.3}} & \underline{\textbf{86.8}}  &  \underline{\textbf{86.6}}   \\

\bottomrule
\end{tabular}
}
% \vspace{-0.1in}
\caption{Results on the CoNLL2005 and CoNLL2012 datasets. ALC-\{Span,BIO\}-ELMo are from \cite{Ouchi2018ASS} and Pred-Span-ELMo is from \cite{He2018JointlyPP}. WSJ (Wall Street Journal) is in-domain and Brown is out-of-domain. (ALC=Alternating LSTM+CRF).}
%BERT-\{(S),(L)\} stand for the cased small and large pretrained BERT embeddings.
\label{tab:allres}
% \vspace{-1.5em}
\end{table*}

The updated base model after the second phase of training is the final SRL model, which we call the Parameterized Neighborhood Memory Adaptive (PNMA) model. 
By parameterizing the representation of $N_K(w)$ and retraining, we allow our model to make optimal use of the nearest neighbors of $w$ in $M$. In contrast to \citet{Akbik2016KSRLIL}, we do not use handcrafted features or distance functions. Compared to \citet{Sprechmann2018MemorybasedPA}, we do not update the classification and CRF layers for each token or sentence separately.

\section{Experiments}

\label{sec:results}

\subsection{Experimental Setup} 
\noindent\textbf{Datasets:} We evaluate PNMA on both span style and dependency style Propbank semantic parsing datasets. For span style SRL evaluation we present results on the CoNLL2005 shared task \cite{Carreras2005} and the CoNLL2012 \cite{Pradhan2012CoNLL2012ST} datasets. 
We evaluate PNMA on the English subset of CoNLL2009 \cite{hajivc2009conll} shared task dataset to present the applicability of PNMA to dependency style SRL datasets. Key statistics of these datasets is in Appendix A.

For the dependency SRL task, we follow the work of \citet{marcheggiani2017encoding} and use the off-the-shelf disambiguator for predicate sense disambiguation \cite{roth2016neural}. 

\noindent\textbf{Word Embeddings:} We experiment with randomly initialized embeddings and publicly available pretrained contextualized embeddings from ELMo and BERT (the large cased model). In both cases, we use a scaled convex combination of embeddings $\mathbf{e}_{i}^{\text{CE}}(w)$ produced by layer $i$ of the underlying language model. 

\noindent\textbf{Training Specifics:}
We use embeddings of size 50 to map the 0/1 predicate  indicators. Each LSTM layer has a hidden dimension of 300. We largely follow the training procedure outlined in \citet{Ouchi2018ASS}. We optimize the loss function with the Adam optimizer \cite{Kingma2015AdamAM} for 100 epochs, with an initial learning rate of 1e-3, which is decreased by half after epoch 50 and 75. A $L2$ weight decay of 1e-4 is used for regularization.

We  apply a dropout of $\delta_l\in\{0.05,0.1,0.15\}$ after each LSTM layer and a dropout of $\delta_e\in\{0.45,0.5,0.55\}$ after the word embedding layer. 
We use 1024 dimensions for randomly initialized word embeddings, the same as ELMo and BERT.
%The base model is identical to the one used in \cite{Ouchi2018ASS}. 
For the memory adaptive training phase, we compute $K=64$ neighbors for each token and train for 20 epochs with a constant learning rate of 4e-4. For both the datasets, we populate a memory $M$ with $15\%$ of the tokens in the respective training sets. We did not observe significant performance differences by increasing memory size beyond this. %We evaluate our models using the official CoNLL SRL evaluation script. 

\subsection{Results}

We evaluate our models using the official CoNLL SRL evaluation scripts. For a span style dataset we use the CoNLL 2005 SRL evaluation script and for a dependency SRL dataset we use the CoNLL 2009 SRL evaluation script. The results are averaged over five runs of model training with random seeds, as summarized in Table \ref{tab:allres}. We show results for the base models trained with randomly initialized word embeddings (Base-Rand in the table) and those with ELMo (Base-ELMo) and BERT (Base-BERT) embeddings and the gains obtained by using PNMA in each case. 

\noindent\textbf{Span SRL Result:} For CoNLL2005, the use of PNMA improves validation and in-domain WSJ test F1 scores for all cases, albeit by varying margins, ranging from 0.1 to 0.6 F1 points. In particular, we observe the highest gains of 0.9, 0.7 and 0.3 F1 points on the out-of-domain Brown test set for the three respective models. The consistent improvements confirm the effectiveness of our proposed PNMA model by exploiting information in the nearest neighbors of tokens in $M$. Coupled with the strong performance of Base-BERT, the Base-BERT+PNMA model improves upon syntax agnostic state-of-art results for CoNLL2005 by margins of \textbf{0.2, 0.3 and 1.8} F1 points on the validation, WSJ and Brown test sets, providing \textbf{the best results among syntax agnostic SRL model}.

Although our model works independent of any kind of syntactic information, we also compare PNMA with models that are syntax aware and find that it is very competitive with syntax aware SRL models. This observation is important since syntax aware models assume the availability of a state-of-the-art syntactic parser which is not always available, and may be challenging to obtain for new domains or languages. 

The results for the CoNLL2012 dataset are also shown in Table \ref{tab:allres}. 
Here again, PNMA results in gains across the board with the best results achieved by Base-BERT+PNMA, improving upon the current state of the art  by \textbf{0.3 and 0.4} F1 points for the validation and test sets. 

\noindent\textbf{Dependency SRL Results:} In Table \ref{table:allres09} we present the PNMA results on the CoNLL2009 dataset and compare it with the state-of-the-art syntax aware and syntax agnostic SRL models. When compared to syntax agnostic models we observe a significant performance gain of $5.2$ absolute F1 points on the out-of-domain Brown set. The results also show that PNMA is very competitive when compared with syntax aware models. Irrespective of whether the model is syntax agnostic or syntax aware, we obtain new state-of-the-art on the out-of-domain Brown set. Here again, we observe the performance gain with the PNMA over the base models. 

\begin{table*}[h!]
\footnotesize
\resizebox{\textwidth}{!}{\begin{tabu}{lcccccc||cccccc}
%\tabucline[2pt]{}
\toprule
  &  \multicolumn{6}{c||}{CoNLL2009}&  \multicolumn{6}{c}{Predicate disambiguation not evaluated} \\  \cmidrule(lr){2-7}\cmidrule(lr){8-13}
  &\multicolumn{3}{c}{Test WSJ}&\multicolumn{3}{c||}{Test Brown} &\multicolumn{3}{c}{Test WSJ}&\multicolumn{3}{c}{Test Brown} \\ \cmidrule(lr){2-4} \cmidrule(lr){5-7} \cmidrule(lr){8-10} \cmidrule(lr){11-13}
  \multicolumn{1}{l}{\multirow{1}{*}{Model}}& \textbf{P} & \textbf{R} & \textbf{F1}& \textbf{P} & \textbf{R} & \textbf{F1}& \textbf{P} & \textbf{R} & \textbf{F1}& \textbf{P} & \textbf{R} & \textbf{F1}\\  \midrule \midrule

 %\hline \hline
% \cmidrule(lr){1-1}\cmidrule(lr){2-4} \cmidrule(lr){5-7} \cmidrule(lr){8-10}\cmidrule(lr){11-13}
\multicolumn{10}{l}{Syntax Aware Models}\\  \cmidrule(lr){1-1} 
\cite{akbik2016k} & \textbf{91.2} & 87.4 &89.3&  {82.1} &77.8 &79.9  &  -  &  -&-&-&  -  &  - \\ 
\cite{li2018unified}+ELmo & {90.8} &88.6 &89.7&  81.0 &78.2& 79.6  &  -  &  -&-&-&  -  &  - \\
\cite{kasai2019syntax} & 89.0 &88.2& 88.6&  78.0 &77.2 &77.6  &  -  &  -&-&-&  -  &  -\\
\cite{kasai2019syntax}+ELMo  & 90.3 &{90.0} &{90.2}&  {81.0} &{80.5} &{80.8}  &  -  &  -&-&-&  -  &  -\\ 
\cite{zhou2019parsing}+ELMo & 89.7 &90.9& 90.3& \textbf{83.9} &\textbf{85.0} &\textbf{84.4}&  -  &  -&-&-&  -  &  -\\ 
\cite{li2019dependency}+ELMo &{90.8} &\textbf{93.5}& \textbf{92.2}& 82.0 &83.4 &82.7&  -  &  -&-&-&  -  &  -\\\midrule \midrule
\multicolumn{10}{l}{Syntax Agnostic Models}\\  \cmidrule(lr){1-1} 
\cite{marcheggiani2017simple}&  88.7 &86.8 &87.7&  79.4 &76.2 &77.7  &  -  &  -&-&-&  -  &  - \\ 
\cite{he2018jointly}& 89.5 &87.9 &88.7&   \textbf{81.7} &76.1 &78.8  &  -  &  -&-&-&  -  &  - \\ 
\cite{cai2018full}& {89.9} &{89.2} &89.6&  79.8 &{78.3}& {79.0}  &  -  &  -&-&-&  -  &  - \\ 
\cite{guan-etal-2019-semantic}+ELMo & \textbf{90.0} &\textbf{89.2} &\textbf{89.6}&  80.0 &\textbf{79.4}& \textbf{79.7}  &  -  &  -&-&-&  -  &  - \\
\multicolumn{10}{l}{Ours}\\  \cmidrule(lr){1-1} 
% Ours & & & & & &\\  \cmidrule(lr){1-1} \textbf{}
Base-Rand  & 86.13& 84.45& 85.28& 75.60& 74.58& 75.09&81.35& 77.24& 79.24& 70.35&64.61&67.36\\ %\hline
Base-Rand+PNMA  &86.05& 84.56& 85.30& 75.66& 74.99& 75.33& 81.26& 77.41& 79.29& 70.48& 65.18& 67.73\\  
Base-ELMo  & 89.73  &  90.13  &  89.93&  82.38  &  83.77&83.07  &  86.91  &  85.55&86.23&80.58&  77.40  &  78.96\\ %\hline
Base-ELMo+PNMA  &{89.91} &  {90.13}  & {90.02}&  {82.51}  &  {83.93}  &  {83.21}&87.17 &  85.56   & 86.36&80.78 &  77.62   & 79.17\\ 
Base-BERT  & \underline{\textbf{90.51}} & 90.88 &90.70&82.99 &86.42  &84.67  &\textbf{88.06} & 86.66& 87.36 &81.53 &81.09&81.31 \\ %\hline
Base-BERT+PNMA &90.03&\underline{\textbf{{91.52}}} &\underline{\textbf{{90.77}}} &\underline{\textbf{83.54}} &  \underline{\textbf{86.45}}&  \underline{\textbf{84.97}}&87.40 &\textbf{87.59} &\textbf{87.50} &\textbf{82.32} &\textbf{81.13} &\textbf{81.72}\\ 
\bottomrule
\end{tabu}}
% \vspace{-0.1in}
\caption{Results on English subset of CoNLL2009 shared task dataset. In each column we bold the corresponding best results and underlined bold numbers represents the best score overall.}
\label{table:allres09}
% \vspace{-1.5em}
\end{table*}

Since we use an off-the-shelf sense disambiguator for all the experiments, we also present the results on argument classification alone in Table \ref{table:allres09} (that is we do not evaluate the predicate sense disambiguation following \cite{Marcheggiani2017EncodingSW})
to show the actual performance gain by PNMA. In all the regimes PNMA outperforms the corresponding base model predictions.

\noindent\textbf{Computation Overhead:} Depending on the size of $M$, the time required to compute $N_K(w)$ may vary. We utilize GPUs to do fast batched computation of distances which results in less than $10\%$ overhead compared to training the base model with BERT. Since the LSTM layers are frozen, the second phase is significantly faster than the first.

\section{Analysis}
\label{sec:analysis}

We analyze factors that may impact the effectiveness of PNMA models. We focus on the models where PNMA alters the base model predictions the most: the Base-ELMo model for CoNLL2005~\footnote{On average Base-ELMo+PNMA alters about 40 samples whereas Base-Bert+PNMA alters only 20 samples} and the Base-Bert model for CoNLL2009~\footnote{On average Base-Bert+PNMA alters about 37 samples whereas Base-ELMo+PNMA alters only 30 samples}. 

\begin{figure}[]
\centering
 \includegraphics[width=0.7\columnwidth]{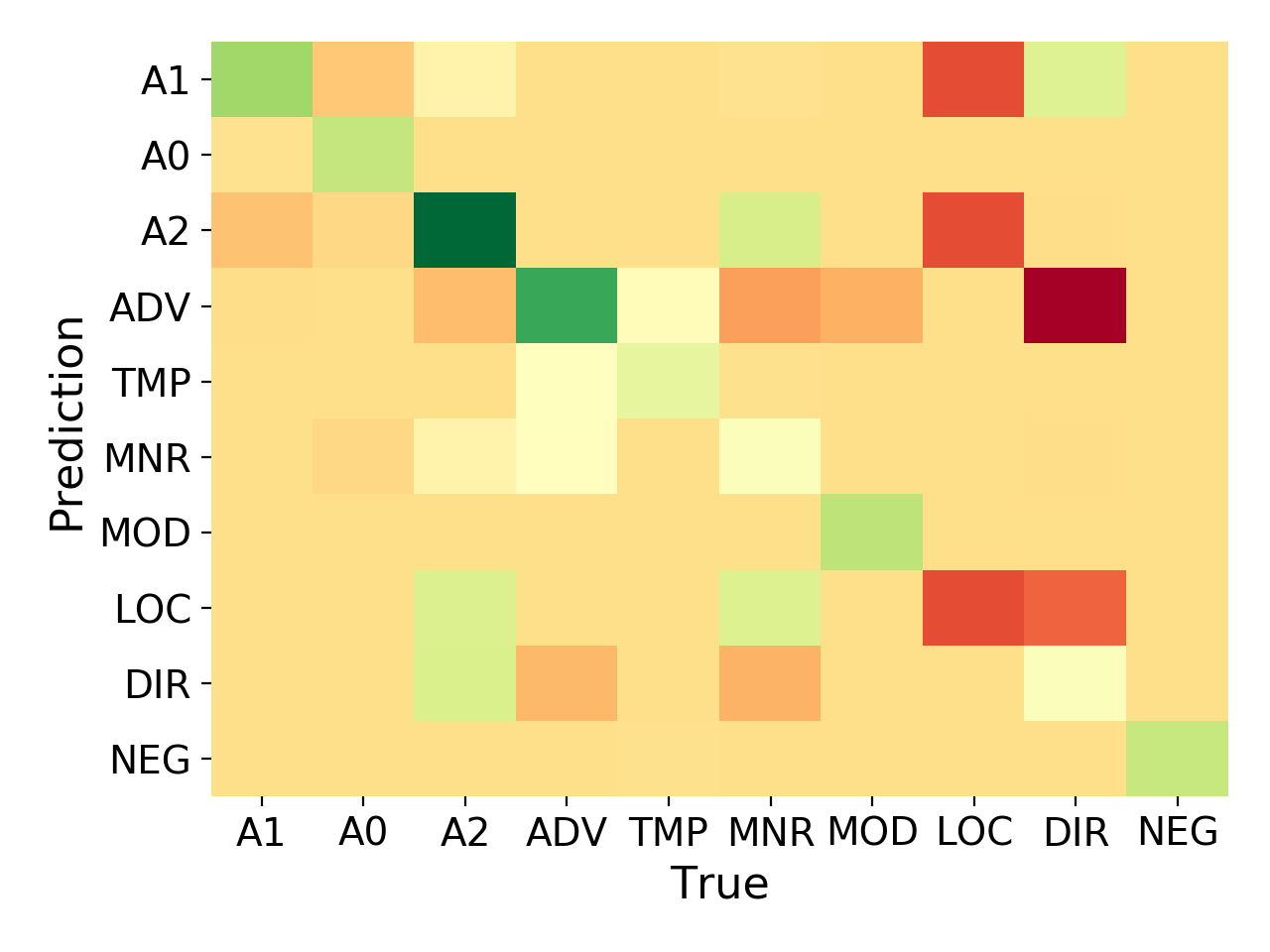}

\caption{Heatmap of the differences between the confusion matrices of Base-ELMo+PNMA and Base-ELMo on the Brown test set in CoNLL2005 for the top 10 argument labels. Greener values on diagonal (more correct labels) and redder values off-diagonal (fewer errors) signal better performance of Base-ELMo+PNMA.}
  \label{fig:pnma_conf}
   % \vspace{-1.5em}
\end{figure}

% From Tables \ref{tab:allres} and \ref{table:allres09}
\subsection{Role Level Analysis}

We first analyze how the effectiveness of PNMA differ across individual argument role labels. Fig.~\ref{fig:pnma_conf} shows a heatmap produced by plotting the difference between the confusion matrices of Base-ELMo+PNMA and Base-ELMo on the Brown test set in CoNLL2005. As can be seen, PNMA improves the prediction of role labels for almost all types in various degrees, except for {\texttt{LOC}}. In particular, PNMA considerably improves the prediction of the core role {{\texttt{A2}}}, mainly by reducing misclassification to {\texttt{A1}} and {\texttt{LOC}}. Similarly, it evidently improves the prediction of {{\texttt{A1}}} by reducing misclassification to {{\texttt{A0}}}, {{\texttt{A2}}} and {{\texttt{LOC}}}. However,  its somewhat poorer performance on the prediction for {{\texttt{LOC}}} indicts slight over-correction of misclassification to {\texttt{LOC}}. Among adjunct roles, {{\texttt{ADV}}} benefits the most from PNMA thanks to reduced confusion with {{\texttt{DIR}}}, {{\texttt{MOD}}} and {{\texttt{A2}}}.
We lists example instances from the CoNLL2005 datasets where PNMA corrects the base model's predictions in Appendix B. % \ref{ap:corr}. 

We further analyze how PNMA leads to better prediction of argument role labels. Taking the core role {\texttt{A2}} as an example. As pointed out in \citet{He2017DeepSR}, {{\texttt{A2}}} shows semantic relations with the adjuncts such as location and direction, often leading to confusion in base model. For instance, for the sentence ``His father would come upstairs and stand at the foot of the bed and look at his son", the span \emph{at the foot of the bed} is classified as \texttt{LOC} by our base model. However, PNMA is able to correctly classify this span with the help of the nearest neighbors (Table \ref{tabel:A2exp}), which contains the valuable information about the correct label and hence improves this confusion over base model.

\newcolumntype{R}{>{\raggedright\arraybackslash}m{11cm}}
\newcolumntype{L}{>{\raggedleft\arraybackslash}m{11cm}}
\newcolumntype{C}{>{\centering\arraybackslash}m{11cm}}

\begin{table}[]
\footnotesize
\resizebox{\columnwidth}{!}{%
\setlength{\tabcolsep}{3pt}\begin{tabular}{cl}
\toprule
Memory Label & Nearest Neighbor\\
\texttt{A2} &..\textcolor{blue}{stood} \textcolor{orange}{next to his car}...\\
\texttt{LOC}& ..\textcolor{blue}{concentrated} \textcolor{orange}{in the northwest}..\\
\texttt{A2} &..\textcolor{blue}{taken} \textcolor{orange}{it for granted}..\\
\texttt{A2} &..\textcolor{blue}{reaped} an extra \$ \textcolor{orange}{on the seventh game}..\\
\texttt{A2} &..\textcolor{blue}{take} skeptics \textcolor{orange}{on a tour of its new Gaylord}..\\
\texttt{LOC} &.. \textcolor{blue}{rising} much more \textcolor{orange}{in the Northeast}..\\
\texttt{A2} &..\textcolor{blue}{fixed} \textcolor{orange}{on the stock market}..\\
\texttt{TMP} &..\textcolor{blue}{dies} \textcolor{orange}{at 49}..\\
\texttt{A2} &..\textcolor{blue}{come} \textcolor{orange}{to a halt}..\\
\texttt{LOC} &.. \textcolor{orange}{in some cases} \textcolor{blue}{become}..\\ \bottomrule
\end{tabular}}
\caption{Top 10 nearest neighbors of the token span \emph{at the foot of the bed}  which is corrected by PNMA from \texttt{LOC} to \texttt{A2}. In each sentence \textcolor{blue}{blue} marks a predicate and \textcolor{orange}{orange} marks an argument span. }\label{tabel:A2exp}
% \vspace{-1.5em}
\end{table}

\subsection{Instance Level Analysis}
For all the datatsets, it is almost four times more likely for PNMA to change a wrong prediction of the base model into a correct prediction than it is for it to change a correct prediction of the base model into a wrong prediction.
Furthermore, focusing on the the samples where PNMA's prediction differs from that of the base model, we find that in  most cases PNMA's prediction is correct: for the CoNLL2005(CoNLL2009) datasets, out of the cases where the models disagree, PNMA is correct in $68\%(48\%)$ whereas the base model is correct in only $20\%(12\%)$ (both are wrong for the rest of these disagreement cases). This shows that the nearest neighbors indeed contain valuable information about the correct label of the test token even when the base model prediction itself is wrong.

\begin{figure*}[h!]
%  \begin{minipage}{0.5\columnwidth}
%  \centering
%     \includegraphics[width=\linewidth]{BA2_from_BA1_16399_beforePNMA.png}
%     \subcaption{Wrong Phase 1}
%     \label{fig::a}
% \end{minipage}%
 \begin{minipage}{0.5\columnwidth}
 \centering
    \includegraphics[width=\linewidth]{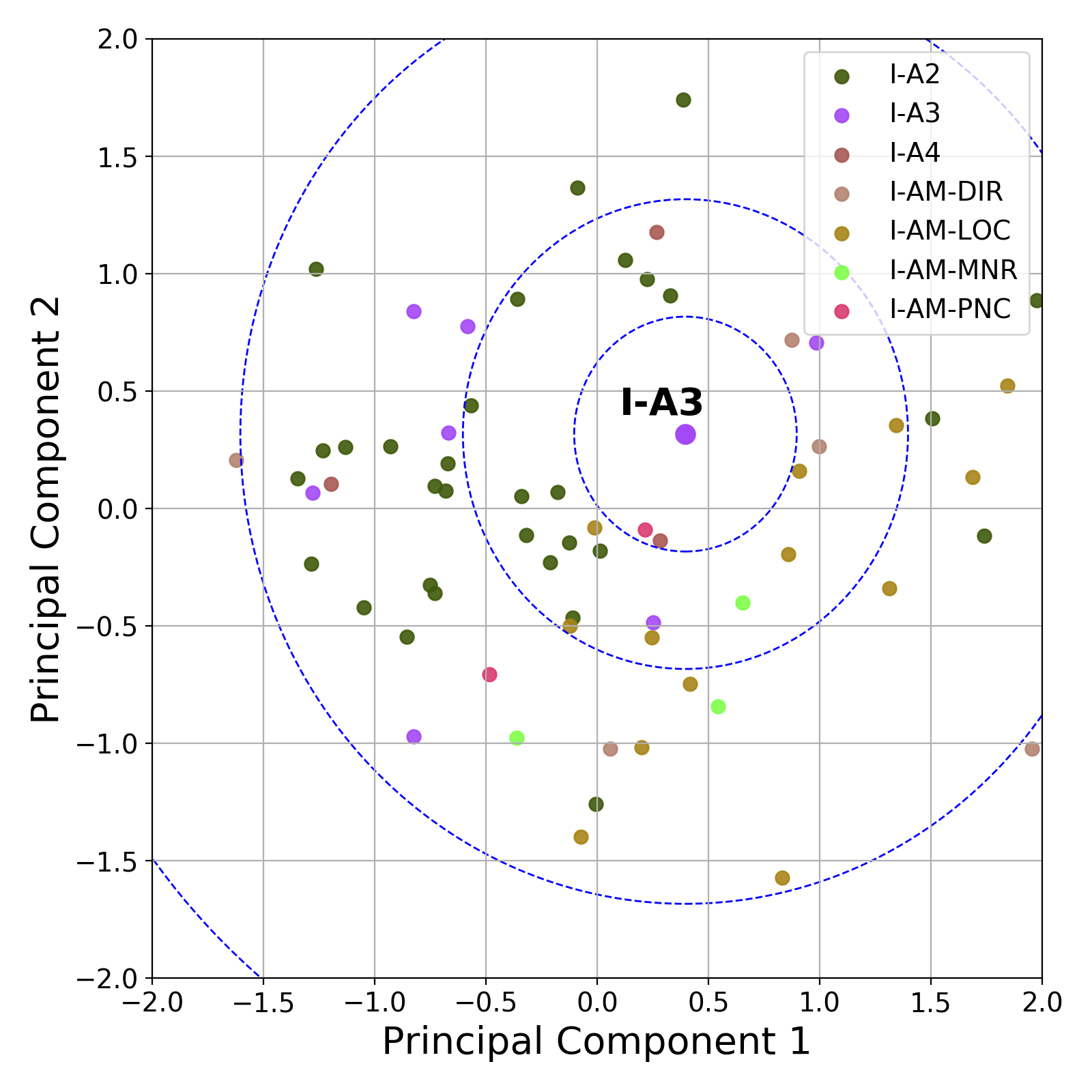}
    % \subcaption{Wrong Base}
    \label{fig::a}
\end{minipage}%
%   \begin{minipage}{0.5\columnwidth}
%     \centering
%     \includegraphics[width=\linewidth]{IA1_from_IA2_14620_beforePNMA.png}
%     \subcaption{Wrong Phase 1}
%     \label{fig::b}
% \end{minipage}%
 \begin{minipage}{0.5\columnwidth}
    \centering
    \includegraphics[width=\linewidth]{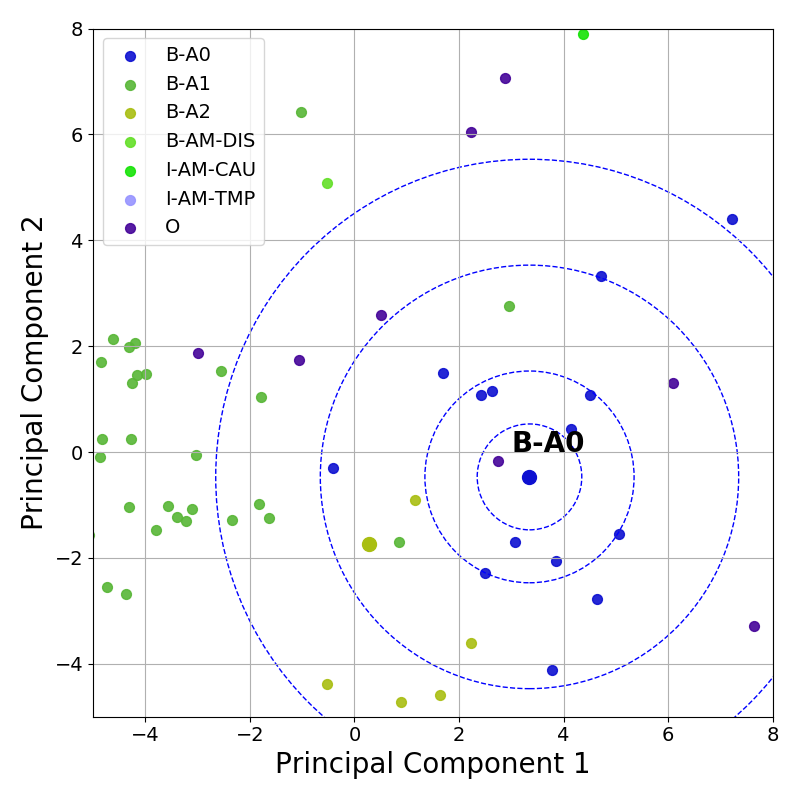}
    % \subcaption{Wrong Base}
    \label{fig::b}
\end{minipage}%
\begin{minipage}{0.5\columnwidth}
    \centering
    \includegraphics[width=1\linewidth]{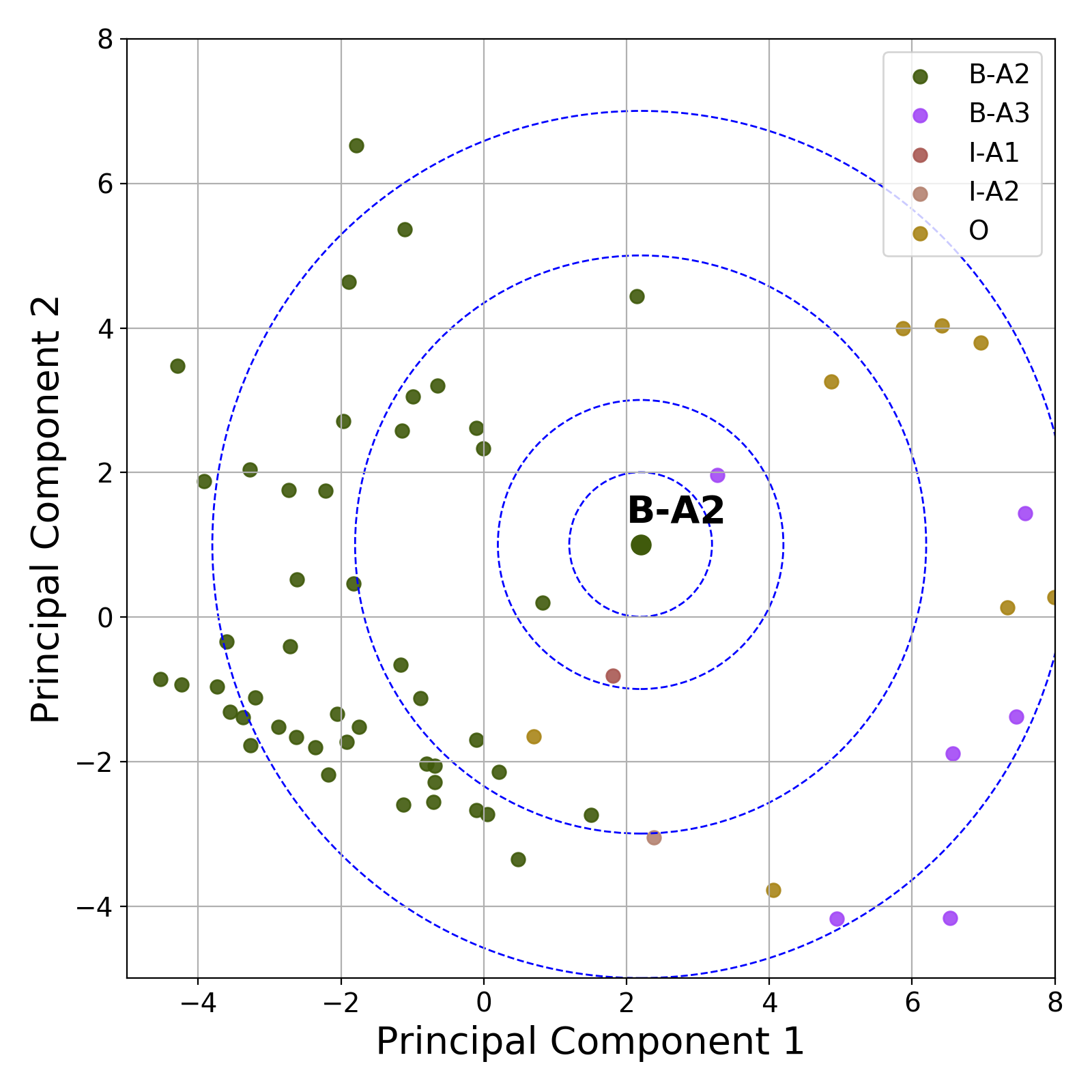}
    % \subcaption{Correct Base}
    \label{fig::c}
\end{minipage}%
\begin{minipage}{0.5\columnwidth}
    \centering
    \includegraphics[width=1\linewidth]{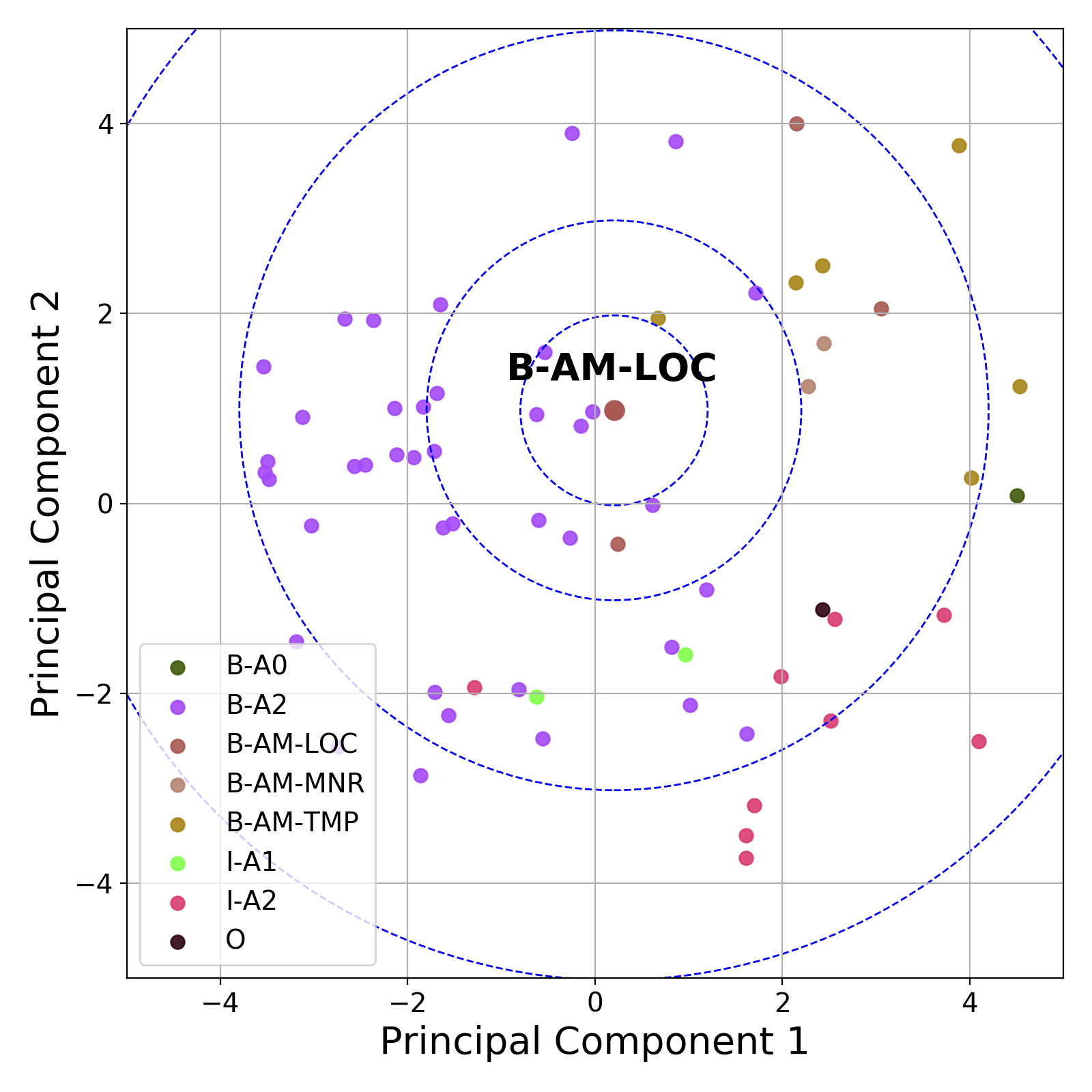}
    % \subcaption{Correct Base}
    \label{fig::d}
\end{minipage}
~
%   \begin{minipage}{0.5\columnwidth}
%   \centering
%     \includegraphics[width=\linewidth]{BA2_from_BA1_16399_afterPNMA.png}
%     \subcaption{\texttt{B-A1} $\rightarrow$ \texttt{B-A2} \\ Correct PNMA\\ (Situation 1)}
%     \label{fig::aPNMA}
% \end{minipage}%
  \begin{minipage}{0.5\columnwidth}
  \centering
    \includegraphics[width=\linewidth]{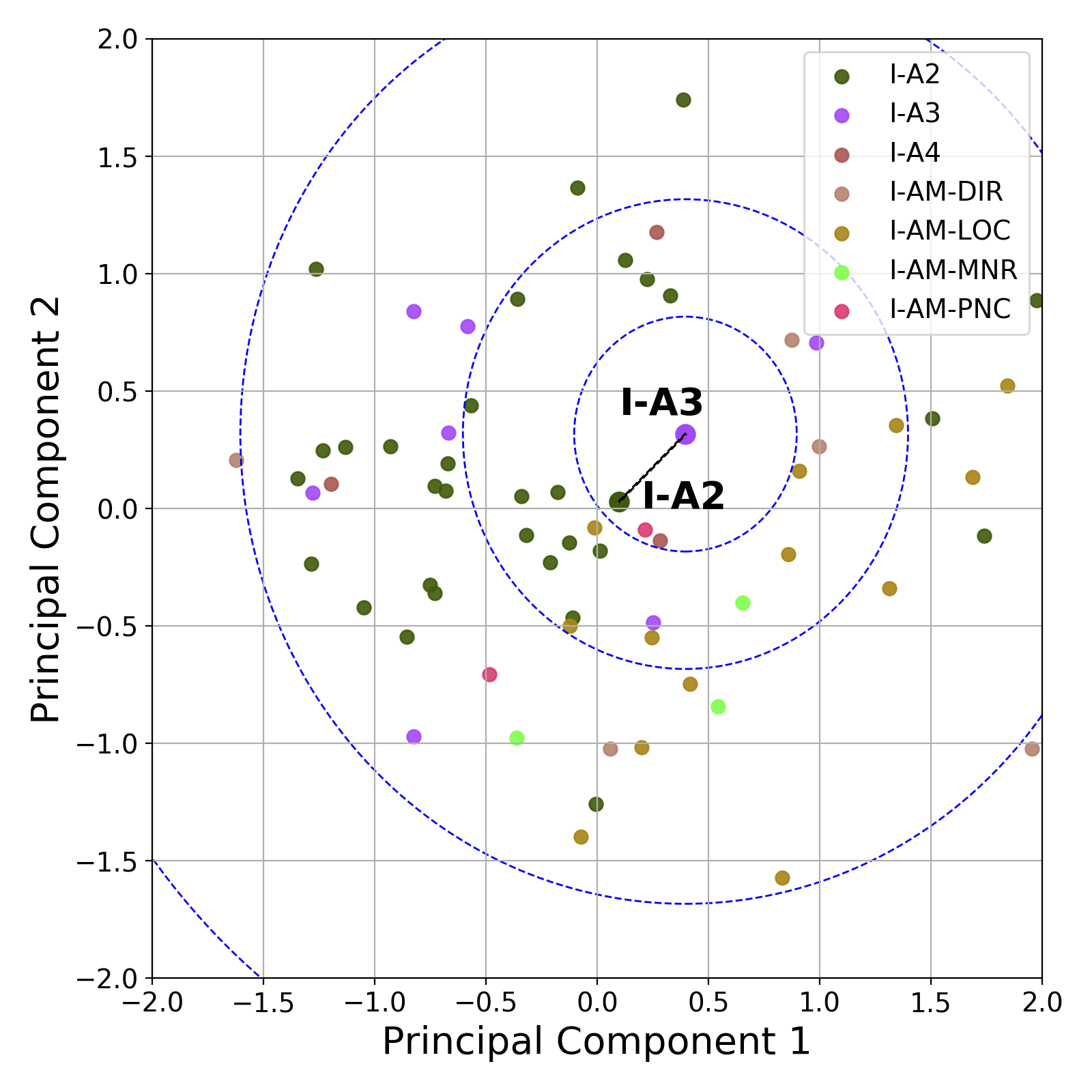}
    \subcaption{Scenario 1}
    % \subcaption{\texttt{I-A3} $\rightarrow$ \texttt{I-A2} \\ Correct PNMA\\ (Situation 1) $2.2\%$}
    \label{fig::aPNMA}
\end{minipage}%
%   \begin{minipage}{0.5\columnwidth}
%     \centering
%     \includegraphics[width=\linewidth]{IA1_from_IA2_14620_afterPNMA.png}
%     \subcaption{\texttt{I-A2} $\rightarrow$ \texttt{I-A1}\\ Correct PNMA\\ (Situation 1)}
%     \label{fig::bPNMA}
% \end{minipage}%
  \begin{minipage}{0.5\columnwidth}
    \centering
    \includegraphics[width=\linewidth]{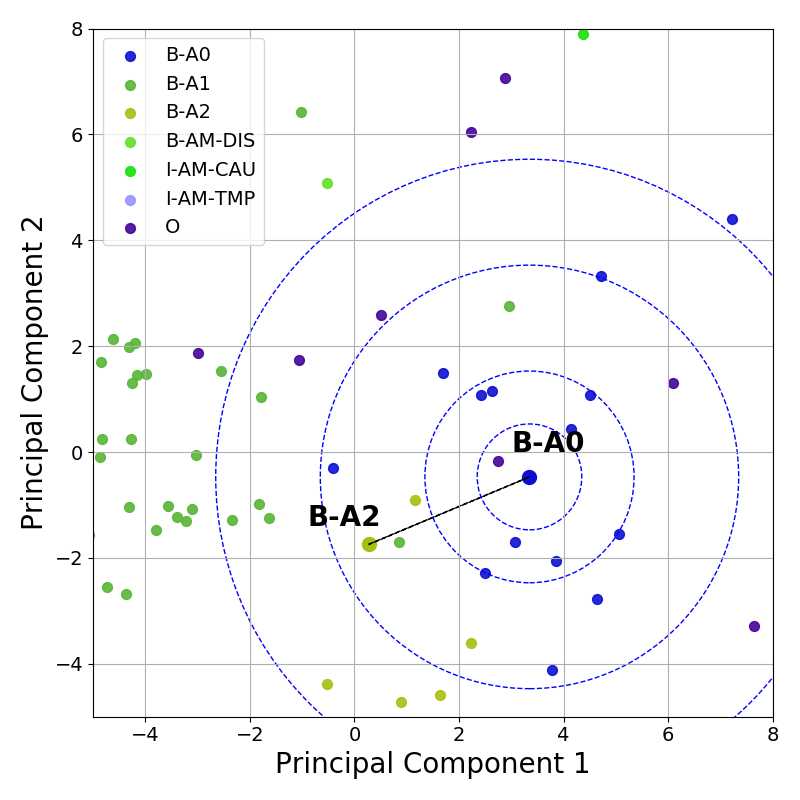}
    \subcaption{Scenario 2}
    % \subcaption{\texttt{B-A0} $\rightarrow$ \texttt{B-A2}\\ Wrong PNMA (B-A4)\\ (Situation 2) $10.0\%$}
    \label{fig::bPNMA}
\end{minipage}%
\begin{minipage}{0.5\columnwidth}
    \centering
    \includegraphics[width=1\linewidth]{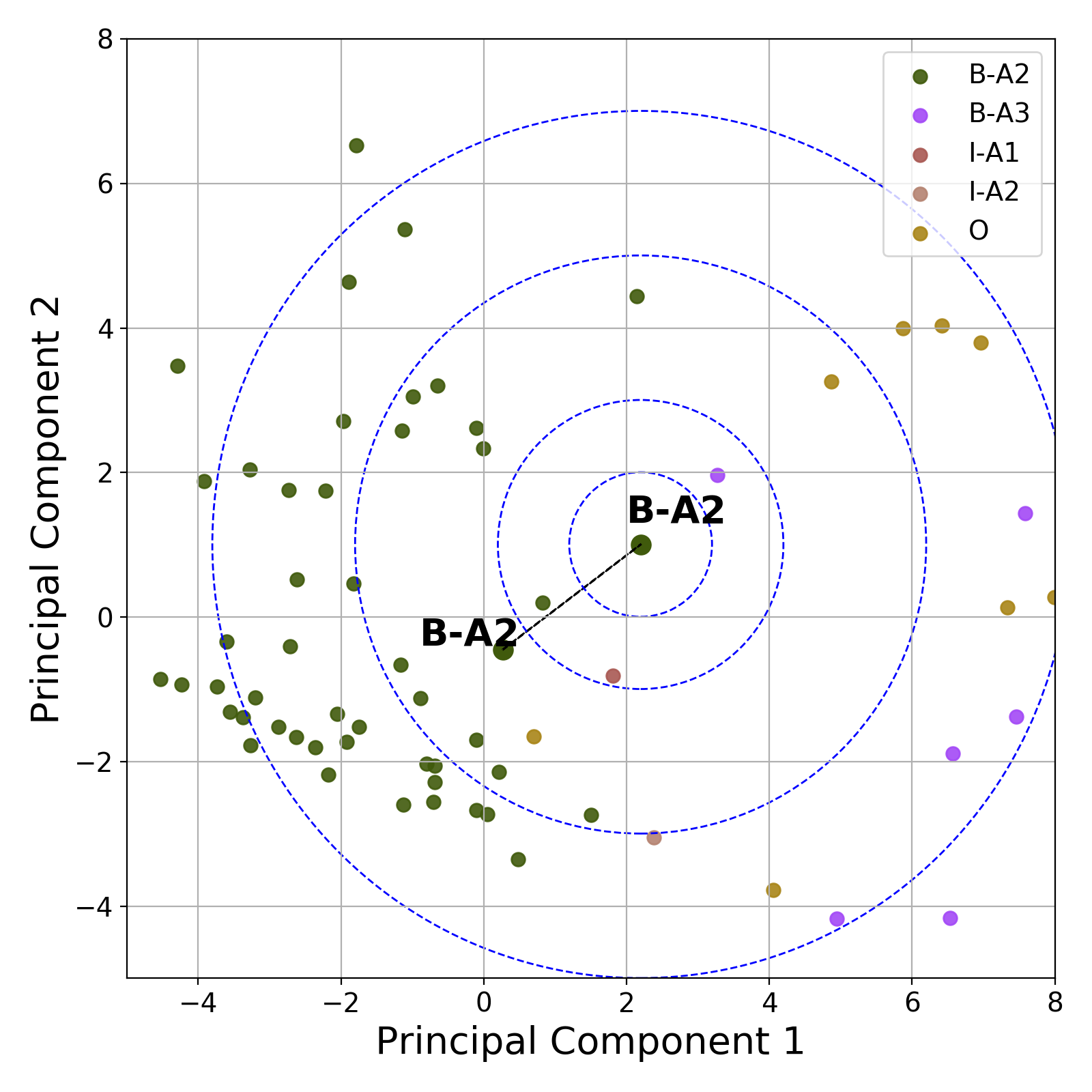}
    \subcaption{Scenario 3}
    % \subcaption{\texttt{B-A2} $\rightarrow$ \texttt{B-A2}\\ Correct PNMA\\ (Situation 3) $86.8\%$}
    \label{fig::cPNMA}
\end{minipage}%
  \begin{minipage}{0.5\columnwidth}
    \begin{center}
    \includegraphics[width=1\linewidth]{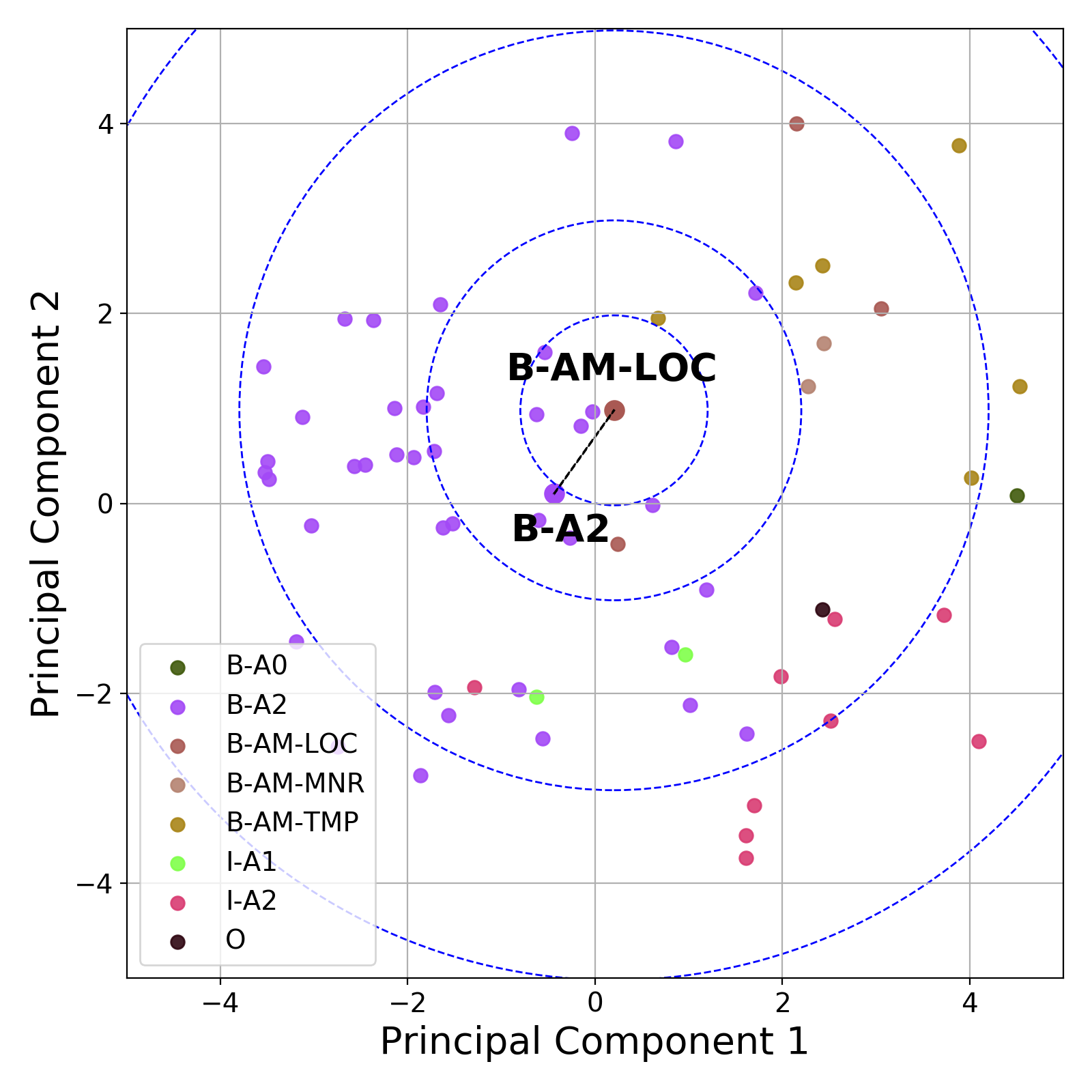}
    \subcaption{Scenario 4}
    % \subcaption{\small{\texttt{B-AM-LOC}} $\rightarrow$ \texttt{B-A2} \\ Wrong PNMA\\ (Situation 4) $0.6\%$}
    \label{fig::dPNMA}
    \end{center}
\end{minipage}%
\caption{Examples of the base model representation of test samples. Each column depicts one test sample. Row 1 depicts the representation of the test sample and the corresponding prediction in the base model; Row 2 depicts the sample representation after PNMA translation and the corresponding change in argument label prediction. The correct labels are: (a) \texttt{I-A2} (b)\texttt{B-A4} (c) \texttt{B-A2} (d) {\texttt{B-AM-LOC}}.}

\label{fig:PNMA}
% \vspace{-1em}
\end{figure*}

Fig. \ref{fig:PNMA} plots the two dimensional representation of a test sample for each scenario, before and after applying PNMA along with the computed nearest neighbors: 1) the base model argument role prediction is wrong; PNMA corrects it in Fig. \ref{fig::aPNMA}; 2) the base model argument role prediction is wrong; PNMA does not alter it in Fig. \ref{fig::bPNMA}; 3) the base model argument role prediction is correct; PNMA fails to alter it in Fig. \ref{fig::cPNMA}; 4) the base model argument role prediction is correct; PNMA alters it to a wrong prediction in Fig. \ref{fig::dPNMA}.

As can be seen in Figs.\ref{fig::aPNMA} and \ref{fig::cPNMA}, PNMA effectively moves the base model's representation of the test sample to an area that is denser in examples with the correct label. In this area, the likelihood of the classifier to predict the correct label is higher. Thus, PNMA enables to correct wrong labels. In Fig. \ref{fig::bPNMA} the correct label is \texttt{B-A4}, but all neighbors are of other labels and hence both models are wrong. There are around $10\%$ of the total samples that lie in this category. Our investigation reveals that most of these samples are not well represented in the memory similar to the situation shown for the example in column (d). That is, for these instances there are a very small number of samples in the neighborhood which have the same label as that of the true label of the current instance.

\begin{figure}[!t]
\centering
\begin{subfigure}{0.47\textwidth}
    \includegraphics[width=1\textwidth]{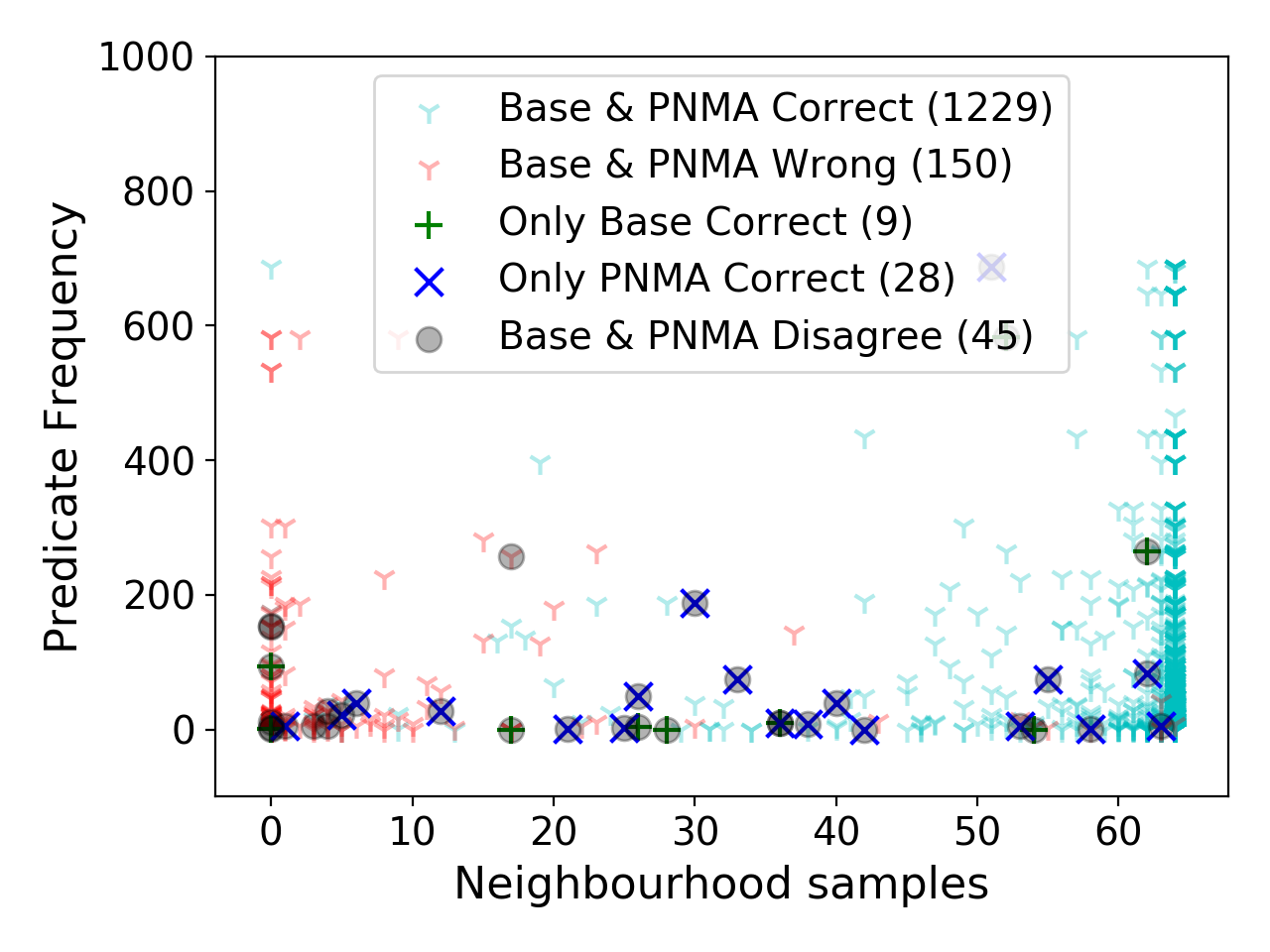}
     \caption{CoNLL2005}\label{fig:neighborVspredfreqa}
    \end{subfigure}
    ~
    \begin{subfigure}{0.47\textwidth}
    \includegraphics[width=\textwidth]{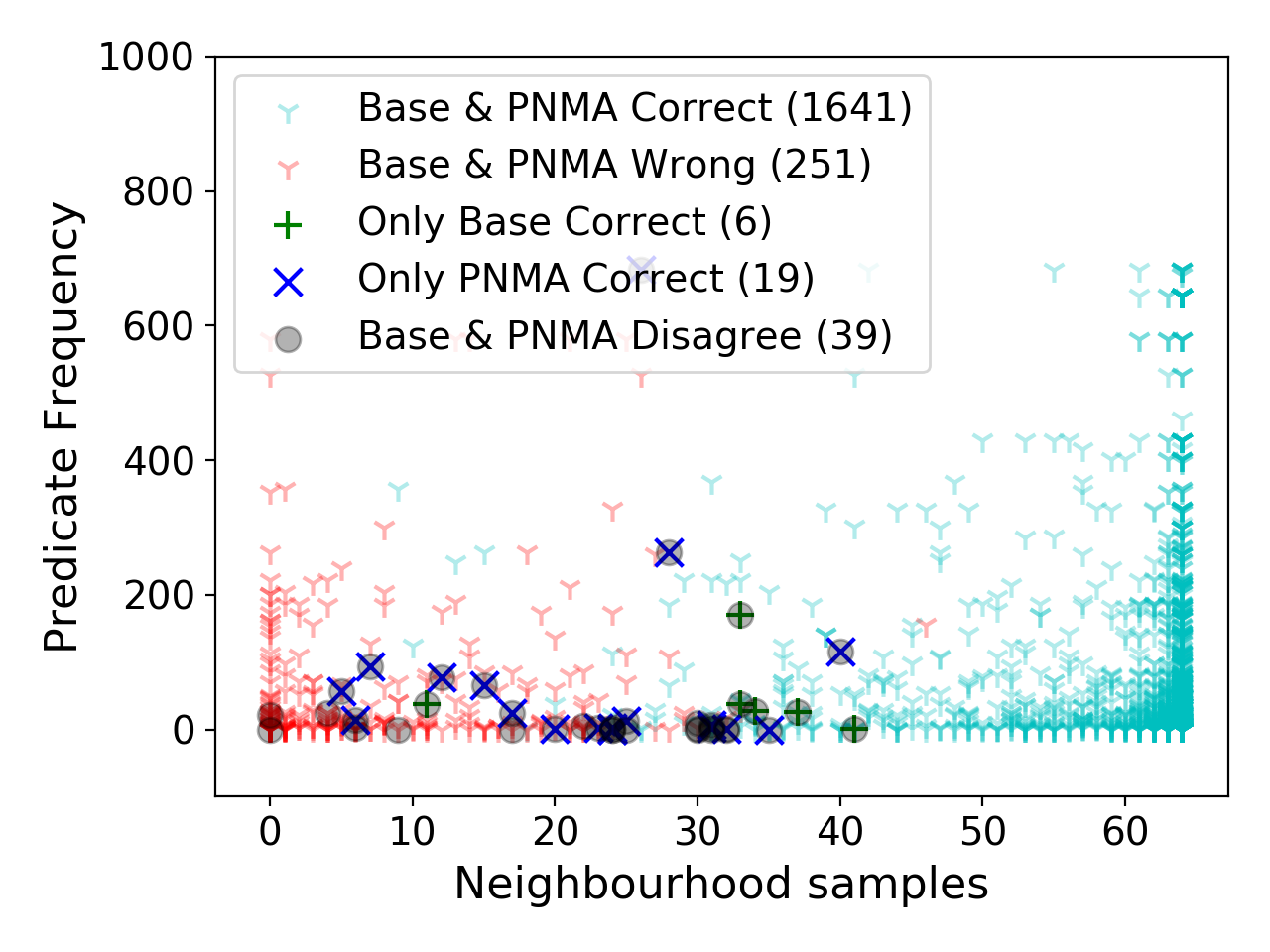}
     \caption{CoNLL2009}\label{fig:neighborVspredfreqb}
    \end{subfigure}
    % \vspace{-0.05in}
    %RA\caption{Correlation between predicate frequency and neighbourhood representation }
    \caption{Model predictions as a function of predicate frequency and neighborhood representation. Neighborhood samples: the number of samples in the neighborhood of the test sample of the same role as the correct role of the test sample.}
    \label{fig:neighborVspredfreq}
    % \vspace{-1.5em}
\end{figure}

\noindent\textbf{Low-Frequency Exceptions:} 
As noted in \citet{Akbik2016KSRLIL}, low-frequency exceptions are important in many real use cases, in which argument roles may be context specific and hence difficult to learn by generalization from the entire training data. Fig.~\ref{fig:neighborVspredfreq} shows how the number of samples on which PNMA disagrees with the base model varies by predicate frequency in the training data. As can be seen, the disagreement between PNMA and the base model is higher in low-frequency predicates. As noted above, when the two models disagree, PNMA is correct most of the time for both CoNLL2005 and CoNLL2009. This observation confirms the effectiveness of PNMA in addressing low-frequency exceptions, similar to instance-learning based K-SRL~\cite{Akbik2016KSRLIL}.

\subsection{Error Analysis}

For PNMA to work, two conditions must be met. First, there should be samples in the neighborhood of the test sample whose gold label matches that of the test sample. Second, enough of those neighboring samples should have representations that results in correct label predictions so as to help PNMA translate the test sample representation to an area which results in a correct label. 

To investigate further, we analyze the predictions of the model as a function of the number of neighborhood samples and the predicate frequency, as shown in Fig. \ref{fig:neighborVspredfreq}. 
These plots show that the PNMA model mostly improves the samples associated with low-frequency predicates and which have moderate representation in the memory, i.e. intermediate number of neighborhood samples (10-40). However, there are some samples that lie in moderate predicate frequency and moderate neighborhood sample regime which are predicted wrong by both models. Although these samples do have a good amount of neighbors that belong to the same category as that of the true label, they are still wrongly predicted by the PNMA model. 
This is because some of these neighbors have a wrong prediction by the base model, which means that the base model representation in the memory is such that it leads to a wrong prediction. When many neighbors have such representations, PNMA will translate the test sample representation into an area which will result in a wrong prediction. This possibly cause the PNMA to misclassify the arguments roles which are correctly classified by the base model. Therefore, to further improve our method, better memory generation techniques are required to cover such cases. This will be explored in future work.

\section{Related Work}
\label{sec:review}
%Semantic role labeling has been studied extensively over the past few years, with deep neural models gaining in popularity due to their superior performance.
Recently, a great attention is paid on the use of deep neural network for the semantic role labeling task \cite{tan2018deep,he2018jointly}. These networks largely fall into two broad categories: \textit{syntax-agnostic} and \textit{syntax-aware}. Syntax-aware models are known to perform better on SRL task \cite{roth2016neural,he2017deep,strubell2018linguistically} and for a long time syntax was considered a prerequisite for better SRL performance~\cite{punyakanok2008importance,gildea2002automatic}. These performance gains are due to the availability of high-quality syntax parser \cite{marcheggiani2017encoding,li2018unified}. However, several works \cite{zhou2015end,marcheggiani2017simple,tan2018deep,he2018jointly} show that the deep networks can extract useful discriminatory features even without the syntactic information.

Span based SRL, formulated as a sequence tagging problem \cite{Villodre2005SemanticRL}, lends itself to end-to-end deep learning models \cite{Zhou2015EndtoendLO}. Using more sophisticated encoders such as LSTMs and Transformers has lead to improved accuracy \cite{He2017DeepSR,Tan2018DeepSR,Zhao2018SyntaxFS,Strubell2018LinguisticallyInformedSF}. An alternate method of directly predicting the spans also achieves comparable results \cite{He2018JointlyPP,Ouchi2018ASS}. In general, context sensitive word embeddings such as ELMo and BERT leads to better performance.
% such as GloVe or Senna \cite{Collobert2011NaturalLP} in many NLP tasks. 

Our model using memory adaptation is inspired by the success of similar techniques in neural models for language modeling e.g. neural cache \cite{Grave2016ImprovingNL,Grave2017UnboundedCM} and memory based methods for few-shot learning \cite{Sprechmann2018MemorybasedPA, Rae2018FastPL}. These methods shown to capture the more long range dependencies. Recently, \cite{guan-etal-2019-semantic} demonstrate the use of memory network in SRL task. Our model is different from \cite{guan-etal-2019-semantic} in the sense that in PNMA nearest neighbors don't need to be labeled and hence can be applied to large unlabeled corpora.

\section{Conclusion}
\label{sec:conc}
To conclude, we propose a Parameterized Neighborhood Memory Adaptive (PNMA) method to exploit information contained in the nearest neighbors of token representations derived from deep LSTM models for SRL. Combined with contextualized word embeddings from BERT, we achieve new state-of-the-art results for single models on both span and dependency style datasets. Our experimental results indicate that PNMA improves over various argument roles and correct most of the sample where PNMA disagree with the base model. For the samples predicted wrongly by both models, we plan to investigate in the future better form of memory or multi-layered representations in the base model to compute better neighborhoods to better handle such cases.

\bibliography{anthology,ref}
\bibliographystyle{acl_natbib}

% \clearpage

\appendix

\begin{table}[ht]
\centering
\resizebox{\columnwidth}{!}{\setlength{\tabcolsep}{3pt}
\begin{tabular}{lllc}
\midrule 
Dataset & Sentences & Tokens & Labels \\
\midrule
CoNLL2005 & 90.8K$|$3.3K$|$5.3K$|$0.8K & 2.6M$|$95K$|$149K$|$19K & 106\\
\midrule
CoNLL2012 & 253K$|$35.3K$|$24.5K & 6.6M$|$935K$|$640K & 129 \\
\midrule
CoNLL2009 &179K$|$6K$|$10K$|$1.2K & 4.5M$|$169K$|$267K$|$26K & 55 \\
\midrule
\end{tabular}}
\caption{Summary of datasets. CoNLL2005 and CoNLL2009 contain an extra out-of-domain test set in addition to the train$|$valid$|$test sets.}
\label{tab:conll_dsets}
% \vspace{-1em}
\end{table}

\section{Dataset}
\label{ap:dataset}
The key statistic of the datasets is shown in Table \ref{tab:conll_dsets}.
For both the span style datasets we convert the semantic predicates and arguments roles to BIO boundary-encoded tags. There are 28 and 42 distinct role types in CoNLL 2005 and CoNLL2012 datasets which were converted to 126 and 129 distinct BIO tags, respectively.
% \FloatBarrier
\begin{table*}[h]
\begin{tabular}{lCc}
\toprule
& & \multicolumn{1}{c}{Base $\rightarrow$ PNMA}\\ \cmidrule(lr){3-3}
\multirow{2}{*}{S1}&    I can {\textcolor{blue}{\textbf{fix}}} \textcolor{red}{\textbf{him}} something later in the afternoon when we get home .          &  \texttt{A1} $\rightarrow$ \texttt{A2}\\
&    I can \textcolor{blue}{\textbf{fix}} him \textcolor{red}{\textbf{something}} later in the afternoon when we get home .     &    \texttt{A2} $\rightarrow$ \texttt{A1}    \\ 
 S2&   The patrol snaked around in back of the cave , \textcolor{blue}{\textbf{approached}} \textcolor{red}{\textbf{it}} from above and dropped in suddenly with wild howls . &   \texttt{A1} $\rightarrow$ \texttt{A2}     \\ 
S3&    “ Will you \textcolor{red}{\textbf{please}} \textcolor{blue}{\textbf{wait}} in here ” .   &   \texttt{AM-MOD} $\rightarrow$ \texttt{AM-ADV} \\ 
S4&    The fear of punishment \textcolor{red}{\textbf{just}} did not \textcolor{blue}{\textbf{bother}} him .     &    \texttt{AM-DIS} $\rightarrow$ \texttt{AM-ADV}    \\ \bottomrule
\end{tabular}
% \caption{Sample instances to illustrate the change in label prediction from Base training to PNMA training. In each sentence the predicate is marked in blue and red denotes the wrong prediction by the base model. }\label{tabel:examples}
\caption{Sample corrections made by PNMA in role classification. In each sentence \textcolor{blue}{blue} marks a predicate and \textcolor{red}{red} marks an argument for role classification. }\label{tabel:examples}
% \vspace{-1em}
\end{table*}
% \FloatBarrier

\section{PNMA Corrections}
\label{ap:corr}
Here we lists the example instances from the CoNLL2005 datasets where PNMA corrects the base model’s predictions. PNMA improves the prediction of role labels for almost all types in various degrees. In particular, PNMA considerably improves the prediction of the core roles.

\end{document}